%% file: main.tex
\pdfoutput=1
\documentclass[11pt,letterpaper]{article}
\usepackage{emnlp2016}
\usepackage{times}
\usepackage{latexsym}

\usepackage[]{algorithm}
\usepackage[]{algorithmic}
\usepackage{times}
\usepackage{latexsym}
\usepackage{makecell}
\usepackage{amsmath}
\usepackage[table,xcdraw]{xcolor}
\usepackage{makecell}
\usepackage{multirow}
\usepackage[table]{xcolor}
\usepackage{hhline}
\usepackage{array}
\usepackage{float}
\usepackage[caption = false]{subfig}
\usepackage[final]{graphicx}
\usepackage{booktabs,caption,fixltx2e}
\usepackage[flushleft]{threeparttable}
\newcolumntype{L}[1]{>{\raggedright\let\newline\\\arraybackslash\hspace{0pt}}m{#1}}
\newcolumntype{C}[1]{>{\centering\let\newline\\\arraybackslash\hspace{0pt}}m{#1}}
\newcolumntype{R}[1]{>{\raggedleft\let\newline\\\arraybackslash\hspace{0pt}}m{#1}}
\newcolumntype{P}[1]{>{\centering\arraybackslash}p{#1}}
\newcolumntype{M}[1]{>{\centering\arraybackslash}m{#1}}
\emnlpfinalcopy

\def\emnlppaperid{***}

\title{Personalized Emphasis Framing for Persuasive Message Generation}
\author{Tao Ding \and Shimei Pan\\
   Department of Information Systems\\
   University of Maryland, Baltimore County\\
  {\tt \{taoding01,shimei\}@umbc.edu}}

\begin{document}
\renewcommand{\baselinestretch}{0.9}
\maketitle

\begin{abstract}
In this paper, we present a study on personalized emphasis framing which can be used to tailor the content of a message to enhance its appeal to different individuals. With this framework, we directly model content selection decisions based on a set of psychologically-motivated domain-independent  personal traits including personality (e.g., extraversion and conscientiousness)  and basic human values (e.g.,  self-transcendence and hedonism). We also demonstrate how the analysis results can be used in automated personalized content selection for persuasive message generation. 
 \end{abstract}

\input{Introduction}
\input{RelatedWorks}

\input{Traits}
\input{PreliminaryStudy}
\input{Experiment}
\input{NLG}
\input{Conclusion}

\bibliography{emnlp2016}
\bibliographystyle{emnlp2016}

\end{document}

%% file: Introduction.tex
\section{Introduction}
Persuasion is an integral part of our personal and professional lives.  The topic of generating persuasive messages has been investigated in different fields with varied focuses. Psychologists focus on the cognitive, social and emotional processes of a persuader and a persuadee to understand what makes a communication persuasive \cite{hovland1953communication,petty1986elaboration,smith1996message}. Marketing researchers are interested in applying theories of persuasion in promoting consumer products \cite{szybillo1973resistance,han1994persuasion,campbell2000consumers,kirmani2004goal,ford2005speak,hirsh2012personalized}.   Natural Language Generation (NLG) researchers are interested in studying the relations between language usage and persuasion in order to build automated systems that produce persuasive messages \cite{guerini2011approaches,reiter2003lessons}.  

It is also generally believed that persuasion is more effective when it is custom-tailored to reflect the interests and concerns of the intended audience \cite{noar2007does,dijkstra2008psychology,hirsh2012personalized}.  A proven tailoring tactic  commonly used by politicians,  marketing executives, as well as public health advocators is content framing \cite{meyerowitz1987effect,maheswaran1990influence,grewal1994moderating,rothman1997shaping}.  Previous framing research has mainly focused on  two types of framing strategies: {\it emphasis frames} and {\it equivalence frames}. To {\it emphasis frame} a message is to simplify reality by focusing on a subset of the aspects of a situation or an issue and make them more salient in a communication  to promote certain definition, causal interpretation and moral evaluation~\cite{entman1993freezing}. For example, in political debating, {\it nuclear energy} can be framed as an {\it economic development} issue,  a {\it safety} issue or an  {\it environmental} issue.  In marketing, the same car can be framed as  a {\it low cost} car,  a {\it  performance} car, or a {\it green} car.  With different framing strategies, the authors try to appeal to individuals with different beliefs and concerns.  In contrast, {\it equivalence framing} focuses on presenting content as either loss-framed or gain-framed messages.  For example, a smoking cessation message can employ a gain-frame like  ``You will live longer if you quit smoking",  or a loss-frame such as ``You will die sooner if you do not quit smoking". Even though the messages are equivalent factually, the frames can influence a receiver's behavior either to encourage a desirable outcome or to avoid an outcome that is unwanted \cite{tversky1981framing}.

In this study, we focus on personalized {\it emphasis framing} which selects a subset of the aspects of an entity (e.g., a car) to enhance its appeal to a given receiver. In a car advertisement scenario, an emphasis framing model  decides which aspects of a car to emphasize to encourage certain user behaviors.  Thus, the communicative goal of emphasis framing is not to simply convey information, but to influence the opinion or behavior of a receiver.  

Using emphasis framing as the framework for personalized content selection, we can take advantage of rich findings in prior framing research that link content selection decisions to a set of psychologically-motivated domain-independent personal traits. This has made our work more generalizable than those relying on application-specific user characteristics (e.g.,  use an individual's smoking habit to tailor a smoke cessation message).   Since content framing is a part of  the content determination process, the model we propose is  a part of the {\it content planner} in a Natural Language Generation (NLG) system~\cite{reiter1997building}. 

There are three main contributions of this work. 

\begin{enumerate}
\item To the best of our knowledge, this is the first effort in building an automated model of emphasis framing for personalized persuasive message generation.
\item We made content selection decisions based on a set of psychologically-motivated application-independent user traits, such as  personality and basic human values, which makes our work more generalizable than the existing works that rely on domain-specific user characteristics and preferences.
\item We propose a cascade content selection model that integrates personalized content selection patterns uncovered in our analysis  in automated persuasive message generation. 
\end{enumerate}

%% file: RelatedWorks.tex
 \section{Related Work}
In the following, we summarize the research that is most relevant to our work including prior psychology and communication studies that link  emphasis framing with personal traits. Since  building computational models of emphasis framing was not the primary goal in these studies, we also include literature on personalized Natural Language Generation. 

\subsection{Emphasis Framing and Personal Traits}
There is a large body of social, marketing and communication theories on framing effects. Tversky and Kahneman \shortcite{tversky1981framing} state that human decisions are controlled partly by the formulation of the problem and partly by the norms, habits, and personal characteristics of the decision-maker. ~\cite{zaller1992nature,zaller1992simple} point out that framing essentially reorganizes information to increase accessibility of an issue dimension by highlighting one cognitive path that had previously been in the dark. Others argue that the framing effect is due to a change in the rank order of the values associated with different aspects through the interaction with the content found within a message \cite{nelson1997toward,chong2007framing,jacoby2000issue}. Although most research agrees that the characteristics of a receiver play an important role in framing effectiveness, there is a significant disagreement in what characteristics of a receiver result in framing effects.  For example, ~\cite{anderson2010framing} states that people with prior attitudes toward an issue can be influenced by frames, while Slothuus \cite{slothuus2008more} and Tabor et al. \cite{taber2009motivated} did not find a framing effect for those with strong values associated with the issue prior to exposure to the frame. The mixed results  may be due to the fact that many of these studies did not take into account that people with different traits  (e.g., different personality) may react to framing strategies differently.

Recently, personalized framing, especially personality-based framing research has become a hot topic. Among them, Hirsh~\shortcite{hirsh2012personalized} investigates whether a persuasive appeal's effectiveness can be increased by aligning message framing with a recipient's personality profile. In this study, for a given mobile phone, they constructed five advertisements, each designed to target one of the five major personality traits. Their results indicate that advertisements were evaluated more positively when they cohered with participants' personality. In a separate study, ~\cite{conroy2012maternal} found that certain personality traits, particularly openness, agreeableness, and conscientiousness mediate framing effects when participants were presented with different frames of political and health issues such as civil liberties, medical treatments, energy, affirmative action, and gun control. 

Inspired by the above research, we also employ psychologically-motivated trait models to capture individual characteristics. In addition to personality, we also incorporate basic human values since framing effects are linked to personal beliefs and motivations.  As a result,  we have significantly increased the scope of our study over prior research. Moreover,  unlike prior research where only messages hand-crafted by experts were used, we are interested in building computational models to automatically select a subset of the aspects based on personal traits. 

\subsection{Personalized NLG}
There is also a large body of work on personalized Natural Language Generation (NLG). For example, STOP is a Natural Language Generation (NLG) system that generates tailored smoking cessation letters based on responses to a four-page smoking questionnaire~\cite{reiter2003lessons}; PERSIVAL customizes the content of search summaries based on its relevance to a given patient's health record~\cite{mckeown2003leveraging};  MATCH is a multimodal dialogue system that tailors the content of its responses based on a user's restaurant preferences;  M-PIRO tailors the words and complexity of museum object descriptions for different audiences (e.g. adults, children, and experts); PERSONAGE~\cite{biewen2011new} and CRAG 2~\cite{gill2012perceptions} vary linguistic styles  to project intended personality in spoken utterances.  Among them, STOP, PERSIVAL and MATCH use domain-specific user models while  M-PIRO, PERSONAGE and GRAG2 employ domain independent user properties, such as expertise and personality. For PERSONAGE and GRAG2,  personality traits are mainly used to adapt linguistic styles. So far, there has not been much work focusing on using domain-independent user traits to automatically adapt message content to improve its persuasive appeal.

%% file: Traits.tex
 \section{Acquiring Personal Traits}
 \begin{figure}
	\centering
    \includegraphics[width=8cm]{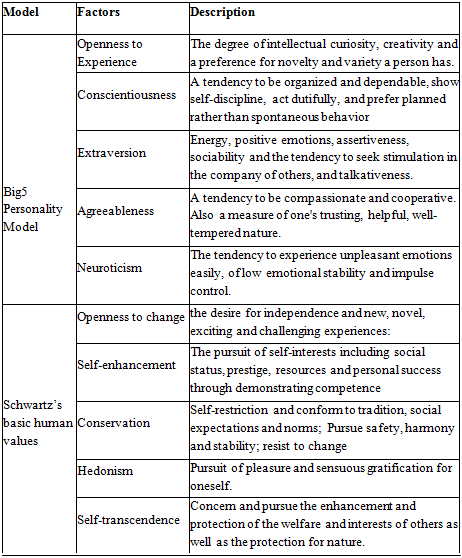}
    \caption{Description of Two Trait Models\label{fig:traits}}
\end{figure}
Since prior study often links framing effects to individual characteristics, such as personality and individual motivations and beliefs, here we focus on two widely-accepted trait models in psychology: the Big5 personality model~\cite{Goldberg:93} and Schwartz's basic human value model~\cite{schwartz:03}.  
%which is often used to explain the motivational bases of attitudes and behavior.  
Figure~\ref{fig:traits} shows the description of each of the Big5 personality traits along with each of the five basic human value traits.
 
To acquire the personality and value traits of a person, traditionally, psychometric tests, such as the IPIP test for Big 5 personality \cite{yarkoni2010abbreviation} and the PVQ survey for values ~\cite{schwartz:03}, were used.  Recent research in the field of psycholinguistics has shown that it is possible to automatically infer personal traits from one's linguistic footprint, such as tweets, Facebook posts and blogs \cite{yarkoni2010personality,celli2013relationships,chen2014understanding}.  Unlike psychometric tests, automated trait analysis allows us to infer personal traits for a large number of people, which makes it possible to scale up automated personal persuasion  for a very large population (e,g., millions of social media users).  
 
 %Goldberg, L. R. (1993). "The structure of phenotypic personality traits". American Psychologist 48 (1): 26?34.
 %Celli et al., 2013] Fabio Celli, Fabio Pianesi, David Stillwell, and Michal Kosinski. Workshop on computational personality recognition (shared task). In Proceedings of the Workshop on Computational Personality Recognition, 2013

%% file: PreliminaryStudy.tex
\section{Acquiring Author Framing Strategy}
Framing effects are often subtle and may be influenced by many factors, such as the characteristics of the authors, the characteristics of the receivers and  the context of the communication. In the first study, we investigate whether it is feasible to build a personalized content selection model based on a writer's (a.k.a. an author's) content framing strategies. 
%To demonstrate the challenge, as a preliminary study, we investigate whether non-expert humans (authors)  are able to customize content for different individuals (receivers) effectively.  The results from this study have also motivated us to build content framing models based on a receiver's traits and preferences in instead of learning from the strategies used by non-expert authors.   

To investigate this, we first randomly generated ten cars, each include eight aspects: {\it safety, fuel economy, quality, style, price, luxury, performance and durability}.  The aspects were selected because similar features were used in prior automobile-related studies~\cite{Atkinson84,Beynon2001}. The value of each aspect was randomly generated on a 5-point Likert scale: ``1 (very bad)", ``2 (bad)", ``3 (average)", ``4 (good)", and ``5 (excellent)".  We also conducted a large-scale personality and basic human value survey on Amazon Mechanical Turk (AMT). We used the  50-item IPIP survey \cite{Goldberg:93} to obtain a Amazon Mechanical Turk worker (a.k.a. Turker)'s personality scores and the 21-item PVQ survey~\cite{schwartz:03} to obtain his/her basic value scores. To ensure the quality of the data from AMT, we added two qualification criteria.  A qualified Turker must  (1) have submitted over 5000 tasks  (2) with an acceptance rate over 95\%. The survey also included several validation questions. The validation questions are pairs of questions that are paraphrases of each other. If the answers to a pair of validation questions are significantly different, the user data were excluded from our analysis. After removing invalid data, we collected the traits of 836 Turkers.  Raw personality scores, ranging from 10 to 50, and raw value scores, ranging from 1 to 6, were computed directly from the survey answers.  The normalized trait scores, ranging from 0 to 1,  were computed using their rank percentiles in this population. 
%For example, a normalized score of 0.75 means its raw score ranks at 75\% in this population (or the raw score is higher than 25\% of the people in this population).

%\begin{table}[h]
%\centering
%\scriptsize
%\caption{An Example of Obtained Traits \label{tbl:persontrait}}

%\begin{tabular}{|c|c|c|c|} \hline
%  Model &Trait &RawS & NormalizedS\\ \hline
% Big& Agreeableness & 22 & 0.981\\ \cline{2-4}
%Five& Conscientiousness & 50 & 0.001\\ \cline{2-4}
%Person & Extraversion &  10 &  1.00\\ \cline{2-4}
%-ality & Neuroticisom & 12 & 0.988\\ \cline{2-4}
%  & openness & 40 & 0.45\\ \hline
%  & Conservation & 3.5 & 0.649\\ \cline{2-4}
 %Basic & Hedonism & 2.0 & 0.933\\ \cline{2-4}
% Human & Openness\_to\_change & 3.0 & 0.897\\ \cline{2-4}
% Values & Self\_Enhancement & 1.5 & 0.989 \\ \cline{2-4}
%  & Self\_Transcendence & 4.417 & 0.717\\ \hline
%\end{tabular}
%\end{table}
%
In addition, we designed two Human Intelligence Tasks (HITs) on AMT:  a content customization task and a validation task. In the content customization task (a.k.a. Task 1), a Turker was asked  to select one car aspect to emphasize in his campaign message for a receiver. The validation task (a.k.a. Task 2) was used to validate whether  a receiver prefers the message customized for her or not.

Specifically, in Task 1,  the Turkers were asked to imagine that they work for a marketing firm on a campaign to promote a new car. Each Turker was given the specification of a car ( randomly selected from the 10 randomly generated cars) and a receiver (randomly selected from the 836 Turkers whose trait scores were known to us).  The Turker was asked to write a campaign message to persuade the receiver to buy the car. But the Turker can only select one of the eight car aspects to include in his message. 
%Since customizing a message based on an interaction of all ten traits can be very challenging, we used a simplified trait profile in our study. The simplified trait profile contains only two traits:  the most prominent personality trait and the most prominent value trait.  The prominence of a trait was defined based on the normalized trait score.  The more different a trait score is from the median (.50), the more prominent the trait is. For example, for the Turker shown in table ~\ref{tbl:persontrait}, the most prominent personality and value traits are ``extraversion" and ``self-enhancement" (highlighted in bold face).  Based on the derived trait profile, the person shown in Table~\ref{tbl:persontrait} is a very introverted person (She has the lowest extraversion score which ranks at 100\% in this population) and does not value achievement and power that much (her self-enhancement score is very low, lower than 98.9\% people in this population).  
For comparison, for the same car, we also asked the same writer to select a car aspect for a different receiver who has an opposite trait profile. The opposite trait profile is defined as the one that is most dissimilar to the given trait profile (with the lowest cosine similarity) in our data.
 %(e.g., for the person shown in table~\ref{tbl:persontrait}, the receiver with the opposite trait profile is someone "very extraverted" and values ``self-enhancement" very much.).  
After the writer selected a car aspect, he also wrote a campaign message using the selected aspect.  Overall, after removing invalid data, we collected 490 pairs of messages for 131 pairs of receivers.  

To validate the framing effect, in Task 2,  we asked a new set of Turkers (receivers) to first complete an IPIP personality survey and a PVQ human value survey. Based on the survey results, we computed the trait profile for each of them. In addition, for each Turker (receiver) in Task 2,  we matched his/her trait profile with the 131 pairs of trait profiles collected in Task 1.  The profile with the highest matching score (computed based on cosine similarity) was selected and its associated message pair was also retrieved.

Then we presented the Turker with a pair of messages, one created for someone with matching trait profile, the other for someone with the opposite trait profile. We randomized the order of the messages.  Finally, we asked the Turker to rate which message they prefer more. If the framing strategies used by the Turkers (authors) in Task 1 were effective, then the Turkers (receivers) in Task 2 will prefer the messages tailored for them more than the ones tailored  for someone with the opposite trait profile.  Overall, after filtering out invalid data, we have collected the results from 145 Turkers.  
%Shimei: Specify the match criteria. What if there is no match.  Were they presented the messages or the features? 
Among the 145 Turkers (receivers) in Task 2, 77 of them prefer the messages written for them while 68  prefer the messages written for someone with the opposite traits. We performed a {\it sign test} to determine whether the difference is statistically significant and the result was negative ($p<0.2$).  

Although moderate personalization effects were found in previous framing research,  only expert-crafted messages were used~\cite{hirsh2012personalized}. Here, when Turkers (mostly non-experts) were asked to customize the messages based on a receiver's traits, no significant effects were found. In the next, since authors' emphasis framing strategies were not effective, we present several experiments designed to automatically derive emphasis framing strategies based on a receiver's traits and his/her aspect selection decisions. 

%% file: Experiment.tex
\section{Learning  Emphasis Framing Strategies}

The goal of this study is to derive emphasis framing patterns based on a receiver's traits and his/her aspect selection decisions.  
\subsection{Data Collection}
We designed another HIT (Task 3) on AMT to collect the data needed for the study. In Task 3, each Turker was asked to take the IPIP and PVQ surveys so that we can obtain their Big5 personality and value scores. In addition, we also asked them to rank all eight car aspects based on their importance to them.  To control the influence of the value of a car aspect on a user's aspect selection decision (e.g.,  if the value of ``safety" is ``poor"  and the value of ``fuel economy" is ``good",  to promote the car, people almost always describe it as "a car with good fuel economy", not  ``an unsafe car", regardless of a receiver's personality). In this study, we kept the values of all car aspects unspecified. After removing the invalid data,  our dataset has 594 valid responses, each contains a Turker's personality and value scores as well as his rank of the eight car aspects.
In the following, we describe how we analyze the relationship between aspect rank and personal traits. 
\subsection{Pattern Discovery with Regression}
\input{tbl/tbl_regression}
In our first study, we employed regression analysis to identify significant correlations between personal traits and aspect ranks. Specifically, we trained eight linear regression models, one for each of the eight car aspects. The dependent variable in each model is the rank of an aspect (from 1 to 8) and the independent variables are the ten user traits.  Here we only focused on the main effects since a full interaction model of ten traits will require more data to train.  Since the raw scores of  the personality and the value traits use different scales, we normalized these scores so that they are all from 0 to 1. Table ~\ref{tbl:regression} shows the regression results. 
Several interesting patterns were discovered in this analysis:  (a)  a positive correlation between the rank of  ``luxury" and ``self-enhancement", a trait often  associated with people who pursue self-interests and value social status, prestige and personal success ($p<0.0001$). This pattern suggests that to promote a car to someone who scores high on ``self-enhancement", we need to highlight the ``luxury" aspect of a car.  (b) the importance rank of ``safety" is positively correlated with ``conservation", a trait associated with people who conform to tradition and pursue safety, harmony, and stability ($p<0.005$). This result suggests that for someone values ``conservation", it is better to emphasize ``car safety" in a personalized sales message. (c)  ``self-transcendence", a trait often associated with people who pursue the protection of the welfare of others and the nature, is positively correlated with the rank of ``fuel economy" ($p<0.005$) but negatively correlated with the rank of ``style" ($p<0.005$). This suggests that for someone who values ``self-transcendence", it is better to emphasize ``fuel economy", but not so much on ``style".  Other significant correlations uncovered in this analysis include a negative correlation between car ``price" and ``conservation" ($p<0.005$), a negative correlation between car ``safety" and ``conscientiousness" ($p<0.05$), and a positive correlation between ``openness to change" and car ``performance" ($p<0.05$). 
%The formula is given as: \\
%\begin{align}
%x^{'}= \frac{x - min(x)}{max(x) - min(x)}
%\end{align} 
%So far,  we have only considered the main effects in the regression analysis. In the following, we further explore methods that allow us to uncover interesting high-order interaction patterns between different traits. 
%In these models, we take main effects into consideration without interactions. We only find one significant correlation in 5 personality, 7 significant correlation in 5 values. The personality model is mostly widely used model to describe how a person engages with the world. When people consider to purchase a car, comparing with their personality, their own values drive them to make the decision. That conscientiousness is negatively correlated with preference of safety ($p=0.02$) which means the conscientious people do not care the safety when they purchase a car. The conservative people tend to like safety ($p=0.002$), and do not care price ($p=0.002$). And for people who open to change, they do not care price ($p=0.04$), but they tend to like performance. The people who desire to self-enhance tend to like luxury car ($p<0.0001$). The people who desire to self transcend do not care about style ($p=0.001$), but concern about fuel economy ($p=0.002$).
\subsection{Pattern Discovery with Constrained Clustering} 
\input{tbl/tbl_cluster_pattern_new} 
%\begin{figure}
%	\centering
%    \includegraphics[width = 7.5cm]{img/feature_preference_level.jpg}
%    \caption{Distributions of Aspect Preferences}
%    \label{fig:level_perference}
%\end{figure}
In the regression analysis, we have only considered the main effects. In order to discover high-order interaction patterns with limited data, we used clustering to group people who share similar traits together. In addition, we also wanted the people within a cluster to share similar aspect preferences.  Otherwise, we won't be able to link the trait patterns discovered in a cluster with aspect preferences. Thus, we employed constrained clustering in this analysis. With constrained clustering, we can ensure the homogeneity of the aspect preferences within each resulting cluster. 
To facilitate this analysis, first we  mapped the aspect ranks obtained in Task 3 into discrete categories. Specifically,  for a complete rank of eight car aspects,  we mapped the top three ranked aspects to an ``Important" class, bottom three to a ``Not-Important" class, and the middle two to a ``Neutral" class. 
%Figure~\ref{fig:level_perference} shows the distribution of these categories for each car aspect. 
%Moreover, for constrained clustering, since it is a semi-supervised algorithm, we will use all the data collected in task 1, 2, and 3. In total, 1475 data points for training and 100 for testing. Among the 1475 training instances, 494 with car aspect preference labels and 981 only with personal traits information. 
In addition, we encoded the aspect homogeneity requirement as additional constraints. Typically, constrained clustering incorporates either a set of must-link constraints, cannot-link constraints, or both. Both a must-link and a cannot-link constraint define a relationship between two data instances. A must-link constraint is used to specify that the two instances in the must-link relation should be placed in the same cluster. A cannot-link constraint is used to specify that the two instances in the cannot-link relation should not be put in the same cluster. These constraints act as a guide for which a constrained clustering algorithm will use to find clusters that satisfy these constraints.  
To encode the homogeneity requirement, for each car aspect (e.g. car safety), we can simply add must-links between every pair of Turkers if they share the same aspect preference (e.g., both consider ``safety" important) and add cannot-links for every pair of Turkers who do not share the same aspect preferences (e.g., one Turker considers ``safety" ``Important", the other considers it ``Not-Important"). But this encoding is not ideal since with both must-links and cannot-links, it is very likely we will get three big clusters for each aspect, each is related to one of the three categories: Important, Neutral and Not-Important. Although the resulting clusters satisfy the homogeneity of the aspect preference requirement, they fail to group people with similar traits together. As a result, in this analysis, we only used cannot-links,  which not only guarantees the homogeneity of aspect preferences, but also creates smaller clusters that group people with similar traits together. 
We employed the Metric Pairwise Constrained KMeans algorithm (MPCK-MEANS)~\cite{bilenko2004integrating} to incorporate the aspect preference homogeneity requirement.  The optimal cluster number $K$ for each aspect was determined empirically by running MPCK-MEANS with different $K$s, $K\in[3,20]$  (3 is the minimum number of clusters since we have 3 different aspect preference categories).  
To determine whether the resulting clusters capture any significant patterns, we used two pattern selection criteria (a) a homogeneity criterion which requires that there is at least one trait whose values in the cluster is relatively homogeneous; (b) a distinctiveness criterion which requires that  for the traits identified in (a), their cluster means need to be significantly different from the population means.  For (a), we used the coefficient of variation (CV) as the homogeneity measure. CV, also known as relative standard deviation (RSD), is a standardized measure of dispersion of a probability or count distribution. It is often expressed as a percentage and is defined as the ratio of the standard deviation $\sigma$  to the mean $|\mu|$. In the study, we required that all the CVs of homogeneous traits to be lower than 0.12.  For (b) we required that the differences of the means need to be significant based on an independent sample t-test with $p<0.001$ and the difference of means is greater than 0.2. Table~\ref{tbl:pattern} highlights some of the patterns discovered using this approach. In this table, we list the cluster id,  cluster label (Important, Not-Important, Neutral), clustering accuracy, and significant traits in the cluster (``+" indicates that the cluster mean is higher than population mean, ``-" means the opposite). For example, based on the Safety-1 pattern, people who are more extraverted (extrave (+)) and more neurotic (neurotic (+)) tend to consider ``car safety" important.   Similarly, based on pattern Safety-3, people who are more conscientious (conscie(+)) but less open (open(-)) tend to consider ``safety" important. Other interesting patterns include: people who are less open (open(-)) and do not value hedonism (hedomis (-)) don't consider performance very important (performance-3), and people who are more extraverted (extrave(+)), value hedonism and self-enhancement (hedonis(+), self-en(+))  do not think durability important (durability-1). 

%% file: tbl/tbl_regression.tex
\providecommand{\e}[1]{\ensuremath{\times 10^{#1}}}
\begin{table*}[ht!]
\captionsetup{font=small}
	\caption{Results of the Regression Analysis \label{tbl:regression}}
%\scriptsize
%\centering
	\makebox[\linewidth]{
		\scriptsize 
		\begin{tabular}{|c|l|l|l|l|l|l|l|l|} \hline
 		&Safety& Fuel & Quality & Style & Price &Luxury & Perf & Durab\\\hline
		Agreeableness & 0.39 & -0.52  &-0.53  &0.54  &0.81 & 0.004& -0.62&-0.27\\\hline
		Conscientiousness & -1.75 *	& -0.31	&0.80  &0.29  &-0.01&0.27&0.83&-0.12\\\hline
		Extroversion	& 0.69&-0.71  &0.008  &-0.25&-0.37&0.48&-0.07&0.224\\\hline
		Neurotisim&1.08 &-0.01&-0.46  &-0.11  &-0.32&-0.07&0.18&-0.28\\\hline
		Openness&	1.59	&-0.05&0.01 &-0.99 &0.36&-0.53&-0.46&0.07	\\\hline
		Conservation&1.99 ** &-0.99 &-0.66&0.84&-1.72 ** &0.21&0.38&-0.03\\\hline
		Hedonism &1.47 & -0.15&-0.69&0.16&0.51&-0.06&-0.82&-0.43\\\hline
		Openness to change&-2.15&0.08&0.58&0.48&-1.99 *&-0.38&2.29*&1.07\\\hline
		Self-enhancement&-1.39&-1.12&0.58&0.47&-0.31&2.41 ***&0.77&-1.41\\\hline
		Self-transcendence&1.33&2.37 **&1.36&-2.47 **&-0.91&-1.01&-0.33&-0.32\\\hline
		\multicolumn{9}{l}{%
  			\begin{minipage}{6.5cm}%
    		\tiny Note: $p < 0.05$, ** $p < 0.005$, *** $p < 0.0001$%
  			\end{minipage}%
		}\\
		\end{tabular}
	}
\end{table*}

%% file: tbl/tbl_cluster_pattern_new.tex
\providecommand{\e}[1]{\ensuremath{\times 10^{#1}}}
\begin{table*}[ht!]
%\centering
\captionsetup{font=small}
\caption{Patterns Discovered in Clustering Analysis \label{tbl:pattern}}
%\scriptsize
\makebox[\linewidth]{
	\scriptsize
	\begin{tabular}{|c|c|c|c|l|} \hline
	Feature & Cluster & Accuracy & Label & Significant traits \\\hline
	\multirow{3}{*}{Safety} 
	%&6& 0.7 &pos&Hedonis(+),Open(+)\\\cline{2-5}
	%&8&0.7 &pos&Hedonis(-),Open(-)\\\cline{2-5}
	&1&0.7 &Important&Extrave(+),Neuroti(+)\\\cline{2-5}
	&2& 0.64&Neutral&Conscie(+),Hedonis(+),Open(+),Self-en(+)\\\cline{2-5}
	&3& 0.71&Important&Conscie(+),Open(-)\\\hline

	\multirow{2}{*}{Fuel}
	&1&0.54 &Neutral&Open(-),Self-en(-)\\\cline{2-5}
	&2&0.54 &Not-Important&Hedonis(+),Open(+),Self-en(+)\\\hline

	\multirow{3}{*}{Quality} 
	&1&0.43&Important&Extrave(+),Neuroti(+)\\\cline{2-5}
	&2&0.45&Non-Important&Hedonis(+),Open(+),Self-en(+)\\\cline{2-5}
	%&14&0.37&pos&Conscie(+),Neuroti(+)\\\cline{2-5}
	&3&0.45&Not-Important&Conscie(+),Open(-)\\\hline

	\multirow{3}{*}{Style}
	&1&0.5&Not-Important&Hedonis(-),Open(-)\\\cline{2-5}
	%&7&0.37&pos&Hedonis(+),Self-en(+)\\\cline{2-5}
	&2&0.55&Neutral&Conscie(+),Extrave(+),Neuroti(+)\\\cline{2-5}
	&3&0.62&Neutral&Conscie(+),Hedonis(+),Open(+),Self-en(+)\\\cline{2-5}
	%&14&0.65&pos&Hedonis(-),Open(-),Self-en(-)\\\cline{2-5}
	&4&0.74&Not-Important&Conscie(+),Open(-)\\\hline

	%Price
	%Luxury
	\multirow{3}{*}{Performance}
	%&10&0.44&neu&Hedonis(+),Open(+)\\\cline{2-5}
	%&13&0.41&pos&Conscie(+),Conserv(+),Hedonis(+),Open(+),Self-en(+)\\\cline{2-5}
	&1&0.73&Neutral&Extrave(+),Neuroti(+)\\\cline{2-5}
	&2&0.5&Neutral&Conscie(+),Open(-)\\\cline{2-5}
	&3&0.4&Not-Important&Hedonis(-),Open(-)\\\hline

	\multirow{2}{*}{Durability} 
	&1&0.56&Not-Important&Extrave(+),Hedonis(+),Self-en(+)\\\cline{2-5}
	%&8&0.37&neg&Conscie(+),Neuroti(+),Open(-)\\\cline{2-5}
	%&10&0.43&neg&Conscie(+),Extrave(+),Neuroti(+)\\\cline{2-5}
	&2&0.36&Important&Conscie(+),Hedonis(+),Open(+),Self-en(+)\\\hline
	
	\multicolumn{5}{l}{%
  			\begin{minipage}{6.5cm}%
    		\tiny Note: $CV < 0.12$ $P < 0.001$ $Diff > 0.2 $%
  			\end{minipage}%
		}\\
	\end{tabular}
}

\end{table*}

%% file: NLG.tex
 \section{Apply Emphasis Framing in NLG}
\begin{figure}
	\centering
	    \includegraphics[width = 7.5cm]{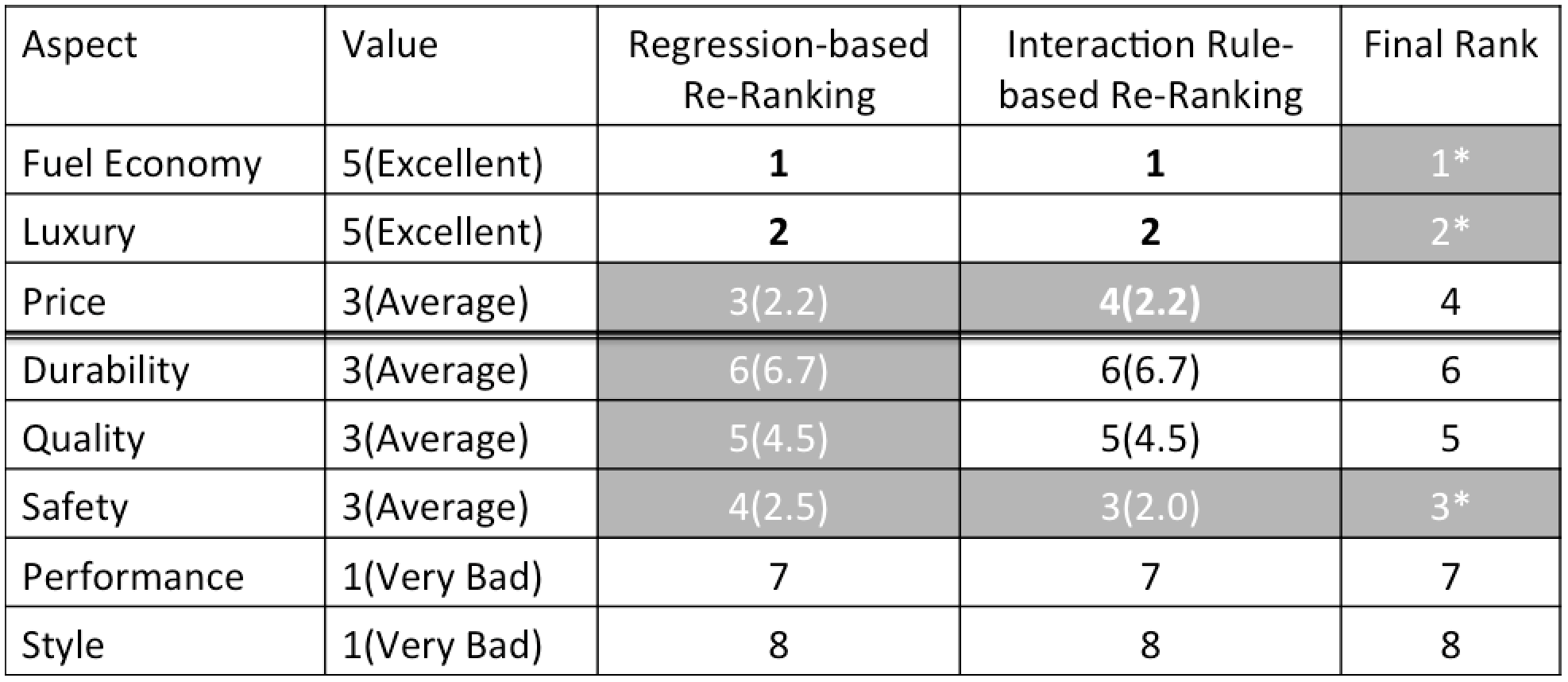}
 \caption{A Cascade Content Selection Example}
\label{fig:nlg}
\end{figure}

The patterns derived in the previous section can be used in personalized content selection for Natural Language Generation. In general, to promote a car, people tend to highlight the good aspects and avoid the bad aspects, regardless of a receiver's personality. For example, people will likely to highlight the fuel economy aspect  if a car is very fuel efficient while de-highlight the same aspect if a car is not fuel efficient. Thus, during content selection, to take the value of an aspect into consideration, we employ a cascade NLG model that integrates value-based content selection with trait-based personalization.  
The input to the cascade content selection model includes: (1) the values of all the aspects; (2) the trait scores of a given receiver; (3) the eight linear-regression models learned in Section 5.2, one for each aspect; (4)  the interaction rules learned in Section 5.3;  (5) n, the number of aspects needed in the output;  (6) the value difference threshold $\delta_1$ that determines whether the values of two or more aspects are significantly different; (7) the rank difference threshold $\delta_2$ that determines whether the ranks of two or more aspects predicted by the linear regression models are significantly different. To select n features to emphasize,  our system first ranks all the aspects based on their values. If the value of the  n-th aspect $v_n$ is significantly better than that of the (n+1)-th aspect $v_{n+1}$ (their difference is greater than $\delta_1$), we output the top n aspects directly.  Otherwise, for all the aspects whose values are either the same or not significantly worse than $v_n$, their ranks will be determined by the trait-based linear regression models. Moreover, after re-ranking relevant aspects based on the predicted ranks from the regression models, if the predicted rank of the n-th aspect ($r_n$) is significantly better than that of the (n+1)-th  aspect $r_{n+1}$ (the rank difference is greater than $\delta_2$), we just output the top n aspects in this list. Otherwise,  for those aspects whose ranks predicted by the linear regression models are not significantly lower than $r_n$, we use the interaction rules discovered in Section 5.3 to adjust their ranking scores (increase the rank by $\delta_2$ if ``Important", decrease by $\delta_2$ if ``Not-Important").  For each aspect, if more than one interaction rule applies, more accurate rules take precedence over less accurate rules.  Finally, the system will output top n aspects in the final list. Figure~\ref{fig:nlg} shows an example.  In this example, we assume n=3, $\delta_1$=1 and $\delta_2$=0.5. We first sorted all the aspects based on their values. Since the values of  ``Fuel Economy" and ``Luxury" are significantly better than the 3rd-ranked aspect ``Price",  their ranks are not affected by personalized aspect selection. Similarly, since the values of  ``Performance" and ``Style" are significantly lower than that of the 3rd-ranked aspect, their ranks are also not affected. Since the value differences among the rest 4 aspects, ``Price", ``Durability", ``Quality" and ``Safety" are all equal or not significant worse than $v_3$, we used trait-based personalized ranks from the regression models to re-rank them (the output ranks from the regression models are shown in the parentheses in the column ``Regression-based Re-Ranking").  After re-ranking these aspects based on the predicted ranks, since the rank of the 3rd-ranked aspect ``Price" (2.2) and that of ``Safety" (2.5) is within $\delta_2$, we use the learned interaction rules to adjust their ranks. Since the predicted ranks of ``Durability" and ``Quality" are much worse than  that of ``Price", their ranks are not affected by the interaction rules.  To apply the interaction rules, assume for a given receiver,  both his extraversion and neuroticism scores are much higher than the population average, the  Safety-1, Quality-1 and Performance-1 rules are applicable.  Since the Safety-1 rule predicts that  ``Safety" is ``Important" to the receiver while none of the rules affects ``Price",   the predicted rank for ``Safety" is increased by 0.5. After this adjustment, the ranks of all the aspects are shown in the ``Final Rank" column.  The top 3 aspects based on the final ranks are selected as the output (marked with a *).  

To evaluate the performance of the cascade content selection model, we conducted an additional AMT study.  Given the specifications of the ten cars in Task1, we asked  each AMT participant to select the top-n aspects to emphasize. Here n=1 and 3. In this task, aspect selection not only depends on the importance of an aspect to a receiver, but also the values of the aspects of a given car. We also acquired the personality and value scores of each Turker based on the IPIP personality and PVQ value survey. Finally, we compared the output of our model  with the aspects selected by the Turkers. We used top-n overlapping percentage as the evaluation metrics. Overall, we collected the aspect selection results from 38 Turkers, each on ten different cars. In total, we collected 380 data instances in our ground truth dataset.  We have tested different $\delta1$ and $\delta2$, the best results were obtained when $\delta1=0$ and $\delta2=0.5$. We compared our model with a baseline which purely relies on the values of aspects to determine their ranks. If two or more aspects have the same value (e.g., the values of both ``Price" and ``Durability" are ``3(Average)", their ranks were determined randomly. Table~\ref{tbl:nlgresults} shows the evaluation results, which demonstrate  a clear advantage of using the proposed cascade content-selection model. The Top-1 agreement is 62\% for the cascade model versus the baseline's 54\%. Similarly, if three aspects are needed in the output, the Top-3 agreement is 87\% versus the baseline's 46\%. All the differences are statistically significant based on the paired-t test ($p\leq0.05$).
%\begin{figure}[b]
%	\centering
	    %\includegraphics[width = 5.5cm]{img/nlgresults.png}
% \caption{Cascade Content Selection Evaluation}
%\label{fig:nlgresults}
%\end{figure}
\begin{table}[h]
\centering
\small
\caption{Cascade Content Selection Evaluation \label{tbl:nlgresults}}
\begin{tabular}{|c|c|c|} \hline
   &Cascade &Baseline\\ \hline
Top-1
agreement	&  0.62& 0.54\\ \hline
Top-3
agreement	&  0.87& 0.46\\ \hline
\end{tabular}
\end{table}

%% file: Conclusion.tex
 \section{Discussion: Domain Generalization}
In general, there are two main challenges in adapting a personalized content selection model trained in one domain to another domain: (1) adapting the {\it data model} from one domain (e.g., restaurant data) to another (e.g., movie data);  (2) adapting a domain-specific {\it user model} (e.g., a user' preferences of restaurant features such as "cuisine type") to a different domain (e.g., a user's preferences of movie features such as "movie genre").  Although our current model is trained in the automobile domain, we adopted a domain-independent user model motivated by psychological theories(e.g., personality and basic human values), instead of using domain-dependent user preference models (e.g, a user's preferences of "fuel economy"). This has made our work much more generalizable than systems that rely on domain-dependent user properties. Moreover, to make our current findings more generalizable, we can apply typical domain adaptation methods such as instance-based~\cite{zadrozny04}  or feature-based transfer learning~\cite{Blitzer06} to further adapt the current results.
  
 \section{Conclusions}
In this study, we analyzed the relationship between an individual's traits and his/her perceived aspect importance. Our analysis has uncovered interesting patterns that can be used to automatically customize a message's content to enhance its appeal to an individual.  We also proposed a cascade content selection model to automatically incorporate the analysis results in automated persuasive message generation. Our evaluation results have demonstrated the effectiveness of this approach.